# Residual Squeeze VGG16

*Hussam Qassim, David Feinzimer, and Abhishek Verma*
Department of Computer Science
California State University
Fullerton, California 92834

Email: {hualkassam, dfeinzimer}(at)csu.fullerton.edu, averma(at)fullerton.edu

**Abstract**

Deep learning has given way to a new era of machine learning, apart from computer vision. Convolutional neural networks have been implemented in image classification, segmentation and object detection. Despite recent advancements, we are still in the very early stages and have yet to settle on best practices for network architecture in terms of deep design, small in size and a short training time. In this work, we propose a very deep neural network comprised of 16 Convolutional layers compressed with the Fire Module adapted from the SQUEEZENET model. We also call for the addition of residual connections to help suppress degradation. This model can be implemented on almost every neural network model with fully incorporated residual learning. This proposed model Residual-Squeeze-VGG16 (ResSquVGG16) trained on the large-scale MIT Places365-Standard scene dataset. In our tests, the model performed with accuracy similar to the pre-trained VGG16 model in Top-1 and Top-5 validation accuracy while also enjoying a 23.86% reduction in training time and an 88.4% reduction in size. In our tests, this model was trained from scratch.

# Introduction

Due to recent advancements in high-performing computing systems, GPUs and large distributed clusters [16] along with the availability of large public image repositories like ImageNet, Deng et al. [17] Convolutional networks have seen a lot of research and development interest as of late (Krizhevsky et al. [2]; Zeiler & Fergus [29]; Sermanet et al. [33]; Simonyan & Zisserman [25]). The ImageNet Large-Scale Visual Recognition Challenge (ILSVRC) [32] has served as a platform for multi-generational large scale image classification systems leading to many advancements in deep visual recognition architectures. ILSVRC has seen everything from high-dimensional shallow feature encodings (Perronnin et al. [11]) (winner of ILSVRC-2011) to deep ConvNets (Krizhevsky et al. [2]) (winner of ILSVRC-2012). Since 2012, deep ConvNets have become a focus of the computer vision field with numerous attempts to improve upon the architecture of Krizhevsky et al. [2]to achieve higher accuracy. Top submissions to ILSVRC-2013 (Zeiler & Fergus [29]; Sermanet et al. [33]) called for a smaller receptive window size and smaller stride in the first convolutional layer. Other areas of improvement have been concerned with the training and testing of dense networks over an entire image and on multiple scales (Sermanet et al. [33]; Howard [1]). Simonyan and Zisserman [25] addressed depth in ConvNet architectural design by adding additional convolutional layers, made possible by the use of very small (3 x 3) convolutional filters in all layers, as shown in the Figure 2. As a result, Simonyan and Zisserman [25] developed a significantly more accurate ConvNet architecture which achieved record-breaking results on ILSVRC classification and localization tasks and similar achievements on other image recognition datasets and tasks such as linear SVM feature classification without the benefit of fine-tuning.

With increased network depth, come more problems stemming from degradation. Degradation begins to saturate network accuracy leading to an expedited failure. Surprisingly, this is not a result of overfitting. Degradation has been shown to be a cause of high training error in [[3], [6]] when network depth was extended with the addition of more layers. Degradation shows that all neural network models aren't easily and equivalently optimized. Residual learning [22] is a recently developed resolution to degradation. In our previous work, we addressed slow convergence, overfitting and degradation by fusing the CNDS network [27] and residual learning connections with shortcuts [22] to build the Residual-CNDS [13]. In [13] we introduce residual connections to the basic CNDS [27] eight-layer structure. Our experiments [13] showed that a combination of both structures enhances the accuracy of the CNDS network [27].

Late research on deep convolutional neural networks (CNNs) focuses on increasing accuracy on computer vision datasets. Multiple CNN architectures exist that attain any given accuracy level. With a given equivalent accuracy, CNN architectures with a smaller number of parameters may have several advantages:

• Deployment on FPGA and embedded systems becomes feasible. Since FPGAs commonly contain 10MB or less of local memory and no remote memory or storage, size is a definite issue. However, a small model can be stored and ran directly on the FPGA rather than being streamed and constrained by bandwidth in real-time [16]. Similarly, on Application-Specific Integrated Circuits (ASICs), a small CNN model can be stored onboard, enabling the ASIC for placement on a smaller die.

• There is less overhead when exporting new models to client devices in production environments. In the field of autonomous driving, companies such as Tesla will often distribute updated models from their servers to customers' cars, a method referred to as an over-the-air update. With this, Consumer Reports has noted that the safety and reliability of Tesla's Autopilot semi-autonomous driving features have seen incremental improvements with recent updates [7]. Unfortunately, these over-the-air updates of current CNN/DNN models may require large data transfers. With larger models, such as AlexNet [2], 240MB of data would need to be sent from the server to the car. A smaller model would require less communication, allowing for more frequent update cycles.

• Compressed models also benefit from more efficient distribution. Communication between servers has limited the scaling of distributed CNN training. In distributed data-parallel

FIGURE 1

FIRE MODULE ARCHITECTURAL, EXPLAINED THE CONVOLUTION FILTERS IN THE FIRE MODULE. THIS
FIGURE DEMONSTRATED, $S_{1x1} = 3$, $E_{1x1} = 4$, AND $E_{3x3} = 4$.
THIS FIGURE ILLUSTRATED THE CONVOLUTION FILTERS WITHOUT THE
ACTIVATIONS. (THE SQUEEZENET PAPER [10] INFLUENCED THE PATTERN OF THIS FIGURE)

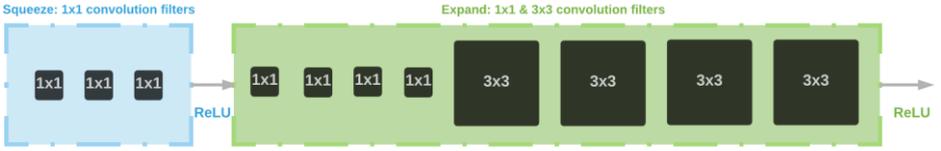

training, communication overhead is directly connected to parameter count in the model [21]. Smaller models would complete a distributed training faster due to this.

Therefore, compressed CNN architectures come with several benefits. This brings us to the task of finding a CNN architecture with a reduced parameter count but accuracy equivalent to that of Simonyan and Zisserman's previous model, VGG16 [26], as shown in Figure 2. We present such an architecture: ResSquVGG16. We also present a further refined approach to searching for novel CNN architectures. This new model brings many advancements, all inherited from its predecessor, VGG16 [26]. Additionally: ResSquVGG16 is smaller and faster than VGG16 [26].

This paper will demonstrate our state of that art technique of adding residual learning to the compressed VGG16 [26]. Our new model also prevents degradation and sees improvements in both time and size. Additionally, the compression method from Iandola et al. [10] improves over the original method in terms of generalization allowing us to declare in confidence that our model can be applied to a wide selection of convolutional neural networks without undergoing any modifications.

Our paper will be organized as follows. Section 1.1 will contain a brief background of the VGG network, residual learning and SqueezeNet. Section 1.2 will contain the details of the proposed: ResSquVGG16 model. Section 1.3 will present details surrounding the large-scale MIT Places365-Standard scene dataset which we used in our experiments. Section 1.4 will present our experimental process. Section 1.5 will contain a discussion of our results and section 1.6 will summarize our work, while the section 1.7 will provide a brief insight into our planned future work.

## 1.1 Background

- **VGG**

    The ConvNet architecture of Simonyan and Zisserman [26] contain several differences from the ones in high-performing entries from the ILSVRC-2012 [3] and ILSVRC-2013 (Zeiler & Fergus [29]; Sermanet et al. [32]) competitions, as you can see the Simonyan and Zisserman [26] model in the Figure 2. First, Simonyan and Zisserman [26] have a very small (3 x 3) receptive field size throughout the entire net, convolved with input at every pixel (with a stride of 1) rather than large receptive fields in the first convolutional layers (e.g. 11 x 11 with a stride of 4 [3] or (7 x 7) with a stride of 2 (Zeiler & Fergus [29]; Sermanet et al. [32])). A stack of two, (3 x 3) convolutional layers

FIGURE 2

THE ARCHITECTURE OF THE VGG16 [26]

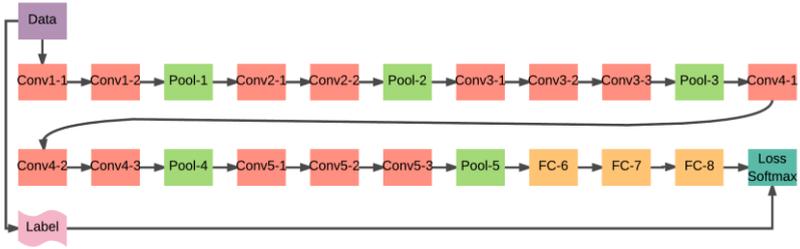

(without any intermixed spatial pooling) has an effective receptive field size of (5 x 5). Three of these layers would have a (7 x 7) effective receptive field. This leads to the question of what Simonyan and Zisserman [26] gained through using a stack of (3 x 3) convolutional layers rather than a single (7 x 7) layer. First, Simonyan and Zisserman [26] used three non-linear rectification layers as opposed to one, rendering the resulting decision more discriminative. Second, parameter count was decreased. Assume both the input and output of a three-layer (3 x 3) convolution stack has C channels. The stack is therefore parameterized by $3(3^2C^2) = 27C^2$ weights, and a single (7 × 7) conv. layer would require $7^2C^2 = 49C^2$ parameters, an increase of 81%. This can be interpreted as imposing a regularization on the (7 × 7) convolutional filters, resulting in a decomposition through the 3 × 3 filters (with non-linearity inserted in between). The use of (1 x 1) convolutional layers increases the nonlinearity of the decision function while avoiding a change to the receptive fields of the convolutional layers. In the Simonyan and Zisserman's [26] model, the rectification function introduces an extra non-linearity even though the (1 x 1) convolution is akin to a linear projection onto the same-sized space. It is important to highlight that (1 x 1) convolutional layers were recently used by Lin et al. [30] in "Network in Network" architecture. Ciresan et al. [5] used small-size convolution filters, however, their nets are much shallower than Simonyan and Zisserman [26], and additionally they did not run any test on the large scale ILSVRC dataset. Interestingly, Goodfellow et al. [14] used deep ConvNets with 11 weight layers to recognize street numbers and their results showed a connection between increased depth and better performance. Another top performer from ILSVRC-2014, GoogLeNet [8], was created independently from Simonyan and Zisserman's [26] work but is similar because it is based on very deep ConvNets (22 weight layers) and small convolution filters. The network topology of GoogLeNet from Szegedy et al. [8] was more complex than that of Simonyan and Zisserman [26], and spatial resolution is reduced in early layers to decrease computation. Nonetheless, Simonyan and Zisserman's [26] model outperforms GoogLeNet [8] in single-network classification accuracy.

- **Residual Learning**

   Where depth should always result in improved accuracy, degradation will decay optimization. Moreover, error in deeper convolutional neural networks is regularly higher when compared to results of superficial neural networks. He et al. [22] has proposed a degradation resolution which allows a portion of stacked layers to approve the current residual mapping where degradation normally stops layers to fit a required subsidiary mapping. This subsidiary mapping will follow formula (2) instead of formula (1). He et al. [22] fount it to be easier to optimize a residual mapping more than a primary mapping.

$$F(x) = H(x) \quad (1)$$
$$F(x) = H(x) - x \quad (2)$$
$$F(x) = H(x) + x \quad (3)$$

In shortcut connections, several layers in a convolutional neural network are skipped [[3],[6],[39]]. Shortcut links are depicted by formula (3) [22]. He et al. [22] use shortcut connections to perform identity mappings. Output sent by shortcut connections is fused with output sent by stacked layers. Shortcut connections have the advantage of remaining parameter free, only attaching trivial numbers for computation. Shortcut connections combined with gating functions with parameters [36] have been coined, Highway networks [35]. Another advantage of He et al. [22] shortcut connections is the option of optimization through stochastic gradient descent (SGD). Identity shortcut connections can be easily implemented through open source deep learning libraries [[1], [28], [34], [40]].

TABLE (1)

RESIDUAL-SQUEEZE-VGG16 ARCHITECTURAL DIMENSIONS (THE PATTERN OF THIS TABLE WAS INFLUENCED BY THE SQUEEZENET PAPER [10])

| Layer name/type | s1x1 (#1x1 squeeze) | e1x1 (#1x1 expand) | e3x3 (#3x3 expand) |
|---|---|---|---|
| Fire1 | 8 | 32 | 32 |
| Fire2 | 16 | 64 | 64 |
| Fire3 | 16 | 64 | 64 |
| Fire4 | 32 | 128 | 128 |
| Fire5 | 32 | 128 | 128 |
| Fire6 | 32 | 128 | 128 |
| Fire7 | 64 | 256 | 256 |
| Fire8 | 64 | 256 | 256 |
| Fire9 | 64 | 256 | 256 |
| Fire10 | 64 | 256 | 256 |
| Fire11 | 64 | 256 | 256 |
| Fire12 | 64 | 256 | 256 |

- **SqueezeNet**

Neural network architectures of deep and convolutional backgrounds often leave space for differing arrangements including choices between micro or macro architectures, solvers and an array of hyperparameters. A healthy amount of research and development has concerned the development of automated processes for generating network architectures with high level accuracy. Some of the more popular processes include Bayesian optimization [18], simulated annealing [38], randomized search [15], and genetic algorithms [24]. These processes achieved improved accuracy when compared against respective baselines. SqueezeNet by Iandola et al. [10] aims to highlight CNN architectures that have a small number of parameters paired with high accuracy. Iandola et al. [1] follow three criteria when generating CNN architectures:

FIGURE 3

THE ARCHITECTURE OF THE RESIDUAL-SQUEEZE-VGG16 WHERE THE FIRE MODULES AND RESIDUAL LEARNING CONNECTIONS WITH SHORTCUTS ARE DEMONSTRATED IN BLUE

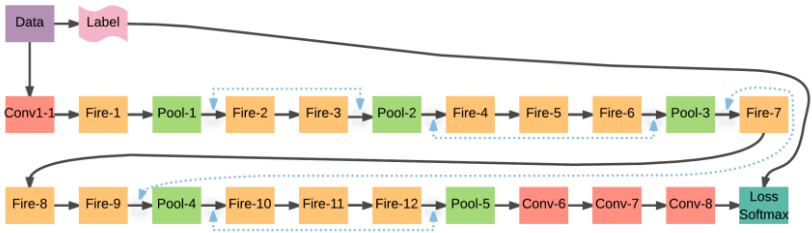

- (1 x 1) filters will take the place of former (3 x 3) filters
- The amount of input channels to (3 x 3) filters will see a reduction
- Downsampling should come later in the network, to provide for large activation maps

Both criteria one and two serve to decrease parameter count while three will ensure maximum accuracy even while working with a limited parameter count.

## 1.2 Architecture

The: ResSquVGG16 contains has twelve Fire Modules [10] and four convolutional layers (Shown in the Figure 3) in comparison to the thirteen convolutional and three fully connected layers of VGG16 [26] (Shown in the Figure 2). A Scale layer will be attached to each Fire Module [10] and the first convolutional layer. The kernel in layer one will be assigned a stride of two and a size of (3 x 3). Next we replace the second convolutional layer with one Fire Module [10]. This is done since the Fire Module [10] has 9x fewer parameters than its (3 x 3) filter equivalent. Input channels are reduced to only (3 x 3) filters. To find the number of parameters in the Fire Module [10] we multiply the number of input channels by the number of filters. To have a CNN architecture with a small parameter count it is important to decrease filter and input channel count. Each Max Pooling layer is assigned a (3 x 3) kernel size following the theory of downsampling late in a network to give convolution layers' large activation maps [16]. Convolutional layers produce activation maps with spatial resolutions of at least (1 x 1) but often much larger. Height and width of activation maps is determined by a set of two factors: size of the input data and the different choices of layers where downsampling will occur. Szegedy et al. [37]; Simonyan & Zisserman [26]; Krizhevsky et al. [2] have all implemented downsampling in CNN architectures by applying a stride greater than one to a selection of convolution or pooling layers. It has been found that when early layers are given large stride parameters, more layers will have small activation maps. It is our belief that large activations maps result in increased classification accuracy. He & Sun [23] observed higher classification accuracy after testing downsampling implemented into four unique CNN architectures.

We use the Fire Module proposed by Iandola et al. [10] in our: ResSquVGG16. This Fire Module [10] is composed of a squeeze convolution layer with (1 x 1) filters fed to an expand layer consisting of (1 x 1) and (3 x 3) convolution filters. Figure 1 depicts a typical Fire Module which contains three tunable dimensions: s1*1, e1*1 and e3*3 [10]. s1*1 [10] represents the count of (1 x 1) filters held in the squeeze layer. e1*1 [10] represents the count of (1 x 1) filters held in the expand layer. e3*3 [10] represents the count of (3 x 3) filters held in the expand layer. We set s1*1 [10] to be less than (e1*1 + e3*3) [10] to limit the count of input channels to the (3 x 3) filters as shown in Table 1.

Formula (12) [22] uses shortcut connections from the residual learning [22] in ResSquVGG16.

$$y = F(x, \{W_i\}) + x \quad (4)$$

After an in-depth review of the Squeeze-VGG16 neural network, it was decided that we'd attach residual learning connections [23] in four locations. Residual connections are attached to locations composed of convolutional layer sequences without any intermediate pooling. Figure 3 shows the described architectures, including residual connections. Element-wise addition links output from Pool1 to output of Fire3. Fire1 has 64 output channels and Fire3 has 128 output channels. We connect a convolutional layer with a 128 size kernel between Pool1 and the element-wise addition layers as a way to equalize the number of output channels.

Next, the second residual connection Pool2 is connected and the shortcut connection exceeds three Fire Module layers. As a result, the residual connection is attached to the output of Pool2 and Fire6. Fire3 has 128 output channels while Fire6 has 256. A new convolution layer with 256 output channels is added following Pool2, preceding the element-wise addition layer to adjust the output channels of Pool2 and Fire6. The third residual connection connects the output of Pool3 and Fire9. Fire6 has 256 output channels and Fire9 512. A new convolutional layer with 512 output channels is added following Pool3, preceding the element-wise addition layer to adjust the output channels of Pool3 and Fire9. Lastly, the final residual connection fuses the output of Pool4 and Fire12. Fire9 and Fire12 both have 512 output channels. No new convolutional layer is added following Pool4 as the number of output channels between Fire9 and Fire12 is already equalized at 512 each.

## 1.3 Image Dataset Description

MIT Computer Science and Artificial Intelligence Laboratory created and maintains the large-scale MIT Places365-Standard [4] dataset. This dataset outsizes ImageNet (ILSVRC 2016) [32] and SUN dataset [20]. There are 1,803,460 total training images in MIT Places365-Standard [4] dataset with 50 validation classes and 900 test classes of sizes ranging from a low of 3,068 to a high of 5,000, as example of the classes and the images shown in the Figure 4. MIT Places365-Standard [4] dataset has classes composed of different scenes which are images labeled with a place or name. MIT Places365-Standard [4] dataset was created to aid in research on computer vision and experiments in this paper were conducted on it.

## 1.4 Experimental Medium and Tactic

During training phase, our proposed network: ResSquVGG16 was trained from scratch: ResSquVGG16 is composed of twelve Fire Modules [10] and four convolution neural layers with four residual connections as opposed to VGG16 [26], which was fine-tuned on the MIT Places365-Standard [10] from pre-trained model on ImageNet ILSVRC-2014 [32] took from the authors of the original paper [26], with thirteen convolutions and three fully-connected layers. To conduct training, we utilize Berkeley Vision and Learning Center's open source deep learning framework, Caffe [40]. We pair Caffe and an open source deep learning GPU

FIGURE 4

SAMPLE OF THE CLASSES AND IMAGES FROM MIT PLACES 365-STANDARD SCENE DATASET [4]

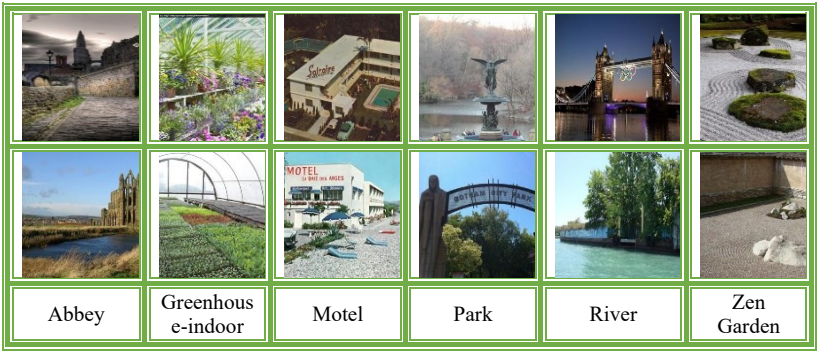

training system, NVIDIA DIGITS [31] allowing users to build and examine artificial neural networks with real-time graphical representations. Our physical hardware consists of four NVIDIA GeForce GTX TITAN X GPUs and two Intel Xeon processors providing us with 48/24 logical/physical cores and a 256GB hard disk. All images in the training and validation datasets are resized to (256 x 256). For preprocessing we perform a subtraction of the average pixel from each color channel of RGB color space. Batch size for training is 128 and validation is 64. This is compared to VGG16 [26] where batch size for training is 256 while validation is 128. The epoch attribute is set to 50 and learning rate to 0.01. Every 10 epochs, the learning rate will degrade 5x and learning average decay will resolve to ½ of the previous value. In VGG16 [26] the epoch attribute is set to 20 and learning rate to 0.001. After the completion of every 4 epochs, the learning rate will degrade 5x and learning average decay will resolve to ½ of the previous value. A randomized crop of size (227 x 227) is applied before introduction to the first convolutional layer. We adapt the weights of all layers from the Xavier distribution with a standard deviation of 0.01 as opposed to VGG16 [26] which used a Gaussian distribution with a 0.01 standard deviation for the weights of each layer. The final convolutional layer of: ResSquVGG16 serves as an output layer with a weight adapted from the Gaussian distribution with a 0.01 standard deviation. Reflection is the only process of augmentation that is performed. During training the model converged after two days and nineteen hours with a size of 1.23gb. By comparison, the original VGG16 [26] converged after three days and sixteen hours with a size of 10.6gb. From this we see that the: ResSquVGG16 model sees a 23.86% speed improvement paired with an 88.40% size reduction from VGG16 [26]. We can see from above that our ResSquVGG16 model surpasses the Simonyan and Zisserman VGG16 [26] in, training from scratch while the VGG16 [25] cannot trained from scratch even in the original paper, ResSquVGG16 model has very less training time if we taking into consideration the batch size (for training is 128 and validation is 64) while VGG16 [26] used (for training is 256 and validation is 128); furthermore, our ResSquVGG16 model trained with 50 epochs while the VGG16 [26] trained with 20 epochs. Find the results of the experiment described above in Table 2 which provides Top-1 and Top-5 accuracy in validation classification.

## 1.5 Results and Discussion

Our paper has brought together the three concepts of VGG16 [26], residual learning [22] and the Squeeze technique [10] to determine if residual connections can enhance VGG16

[26] accuracy, size and speed. We modified the Fire Module concept [10] and implemented our method for determining when and where residual connections should be added. The network does not see a significant increase in complexity following a small amount of computations for the collection process due to residual connections being parameter free. Furthermore, our network size, training time and complexity was reduced with the help of

TABLE (2)

COMPARISON OF THE TOP 1 AND TOP 5 VALIDATION SEGMENTATION ACCURACY (%), DURATION, SIZE, AND NUMBER OF EPOCHS BETWEEN THE FINE-TUNED-VGG16 [26] AND OUR RESSQUVGG16 ON THE MIT PLACES 365-STANDARD DATASET [4]

| Network | Top-1 Validation | Top-5 Validation | Duration | Size | No. of Epoch |
|---|---|---|---|---|---|
| Fine-Tuned-VGG16 | 54 | 84.3 | 3 Day 16 Hour | 10.6 GB | 20 |
| ResSquVGG16 (Our) | 51.68 | 82.04 | 2 Day 19 Hour | 1.23 GB | 50 |

the Fire Modules [10] and saw only a marginal Top-1 and Top-5 accuracy loss. Table 2 compares results between our new network: ResSquVGG16 and the original VGG16 [26] on Top-1 outcome based on the MIT Places 365-Standard dataset [4]. Our new network: ResSquVGG16 got a result of 51.68% compared to VGG16's [26] result of 54% (after fine-tuning from ImageNet ILSVRC-2014 [32]), a difference of only 2.32%. Top-5 results were also comparable with: ResSquVGG16 at 82.04% and VGG16 [26] at 84.3%, a difference of only 2.26%. With this said: ResSquVGG16 enjoys a size 88.4% smaller and a speed 23.86% faster in training than pre-trained VGG16 [26].

## 1.6 Conclusion

In this paper we proposed the: ResSquVGG16 network which comes with a modified Fire Module [10] and a novel method of determining when and where to place residual connections [22]. The Fire Module [10] allowed us to reduce the size, complexity and training time of our network while only seeing a very slight loss in Top-1 and Top-5 accuracy. Our experiments were conducted on the MIT Places 365-Standard [4] dataset, which highlights our improvements on size, time and complexity from VGG16 [26] benchmarks.

## 1.7 Future Work

In future work we plan to focus on applying the methods we've outlined in this paper to other well-regarded networks possibly including ResNet [22] and Densely Connected Convolutional Networks [12] and others. In these future models we hope to also see little or no Top-1 and Top-5 accuracy loss with equal or better reductions in size and complexity.